\pdfoutput=1

\documentclass[10pt,twocolumn,letterpaper]{article}

\usepackage{iccv}
\usepackage{times}
\usepackage{epsfig}
\usepackage{graphicx}
\usepackage{amsmath}
\usepackage{amssymb}

\usepackage{multirow}
\usepackage{float}
\usepackage{booktabs}
\usepackage{fontawesome}
\usepackage[accsupp]{axessibility}

\usepackage[breaklinks=true,bookmarks=false]{hyperref}

\iccvfinalcopy 

\ificcvfinal\fi
\begin{document}

\title{DetermiNet: A Large-Scale Diagnostic Dataset for Complex Visually-Grounded Referencing using Determiners}

\author{
Clarence Lee$^{*1}$, M Ganesh Kumar$^{*2}$, Cheston Tan$^{2}$\\
Design and Artificial Intelligence, SUTD $^1$, Centre for Frontier AI Research, A*STAR $^2$\\
{\tt\small clarence\_leesheng@mymail.sutd.edu.sg, \tt\small m\_ganeshkumar@u.nus.edu, \tt\small cheston\_tan@cfar.a-star.edu.sg}
}

\maketitle
\ificcvfinal\thispagestyle{empty}\fi

\def\thefootnote{*}\footnotetext{Equal contribution}
\def\thefootnote{\arabic{footnote}}

\begin{abstract}
    State-of-the-art visual grounding models can achieve high detection accuracy, but they are not designed to distinguish between all objects versus only certain objects of interest. In natural language, in order to specify a particular object or set of objects of interest, humans use determiners such as ``my'', ``either'' and ``those''. Determiners, as an important word class, are a type of schema in natural language about the reference or quantity of the noun. Existing grounded referencing datasets place much less emphasis on determiners, compared to other word classes such as nouns, verbs and adjectives. This makes it difficult to develop models that understand the full variety and complexity of object referencing. Thus, we have developed and released the DetermiNet dataset~\footnote{\url{https://github.com/clarence-lee-sheng/DetermiNet} contains the dataset and code}, which comprises 250,000 synthetically generated images and captions based on 25 determiners. The task is to predict bounding boxes to identify objects of interest, constrained by the semantics of the given determiner. We find that current state-of-the-art visual grounding models do not perform well on the dataset, highlighting the limitations of existing models on reference and quantification tasks.
\end{abstract}

\section{Introduction}
Humans combine visual and linguistic cues to perform object localization, referencing and quantification tasks on a daily basis. For example, when someone says ``pass me a cup'', we first locate any cups present, and then select one cup based on other criterias, such as the nearest or cleanest one. Deep learning models~\cite{bajaj2019g3raphground, cirik2018using, corona2022voxel, gao2022efficient, kamath2021mdetr, krishna2018referring, nottingham2021modular, suglia2021embodied, wang2022ofa, yu2016modeling} can localize object impressively to achieve the first part of the task. However, the ability to deal with a variety of complex referencing and quantification to achieve the second part of the tasks has yet to be properly investigated.

A \textit{determiner} is an English part-of-speech (word class) that quantifies or references the noun following it. For instance, the determiner in ``\underline{my} apple'' versus ``\underline{your} apple'' takes reference from different owners. The number of apples being referenced differs for ``\underline{some} apples'' versus ``\underline{all} apples''. Such semantic differences are succinctly captured by determiners, and not by other word classes. 


Determiners like ``a'', ``the'' and ``my'' are ubiquitous and among the most common English words \cite{oxfordenglishcorpus, leech2014word}. Most children learn to use determiners at a near-mastery level by 3 years of age~\cite{abu2004describing, brown2013first}. Since determiners play an important role in the semantics of a phrase, they are distinctly classified in natural language processing libraries~\cite{loper2002nltk,santorini1990part}.

Unlike numerous nouns, verbs and adjectives, there are only about 50 determiners in the English language \cite{leech2014word}. Nevertheless, determiners can be highly complex, and a hardcoded or fixed-rule approach to using or understanding determiners simply will not work. For instance, take the determiner ``some'' -- in its simplest form, ``some'' refers to a relatively small number or quantity. However, this can be highly noun-specific and context-specific, \eg the absolute physical quantities for ``add \underline{some} salt'' versus ``drink \underline{some} water'' are very different. Furthermore, determiners that describe ownership or possession, such as ``my'' and ``your'', are highly context-dependent and dynamic, as possession can change on the fly, \eg after handing over an object. In general, there are many such subtleties and complexities for determiners. Hence, a learning-based approach is needed, along with suitable training data. 

If state-of-the-art models could learn a schema of determiners~\cite{rumelhart1980building,ganeshkumar2022biologically, rumelhart1977representation}, it could facilitate flexible combination in novel contexts~\cite{kumar2023oneshot,kansky2017schema, mcclelland2013incorporating} and improve visual reasoning. However, existing vision-language models such as CLIP~\cite{radford2021learning} and BLIP-2~\cite{li2023blip} do not capture the semantic organization of determiners well (see Supplementary Material), and there is no visual grounding dataset that focuses on Determiners. Existing grounded referring expression datasets~\cite{antol2015vqa, goyal2017making, johnson2017clevr, kazemzadeh2014referitgame, mao2016generation, shridhar2020alfred, thomason2022language} exclusively focus on ``the'' and ``a'', making an unambiguous reference to a specific single object. Some examples include ``bottle with a lid'', ``the blue truck in the bottom right corner'' and ``a bird that is close to the baby in a pink shirt''. In other words, existing datasets focus on the noun, verb and adjective aspects of referring expressions, with ``the'' and ``a'' as the main determiners used.

Hence, as a first step towards bridging this gap, we developed the \textit{DetermiNet} diagnostic dataset~\cite{johnson2017clevr} to benchmark current state-of-the-art (SOTA) algorithms on their potential for learning determiner concepts. As with CLEVR~\cite{johnson2017clevr}, good performance on DetermiNet is not an end-goal in itself, as knowledge of the dataset generation process can be used to hand-craft toy models that will not generalize to real-world determiner usage. The dataset uses a bounding box localization task, set in a highly-constrained instruction task context, and deals only with simplified determiner definitions. Even with all these simplifications, we find that SOTA methods do not perform well.

DetermiNet contains 250,000 synthetic images and captions covering 25 determiners. The images are designed with the premise of two avatars interacting at a table with objects. The captions consist of a determiner followed by a noun; the task context is that the viewer is asking the avatar in the image to ``pass me \{\textit{determiner noun}\}''.

\begin{figure*}
  \centering 
  \includegraphics[width=\textwidth]{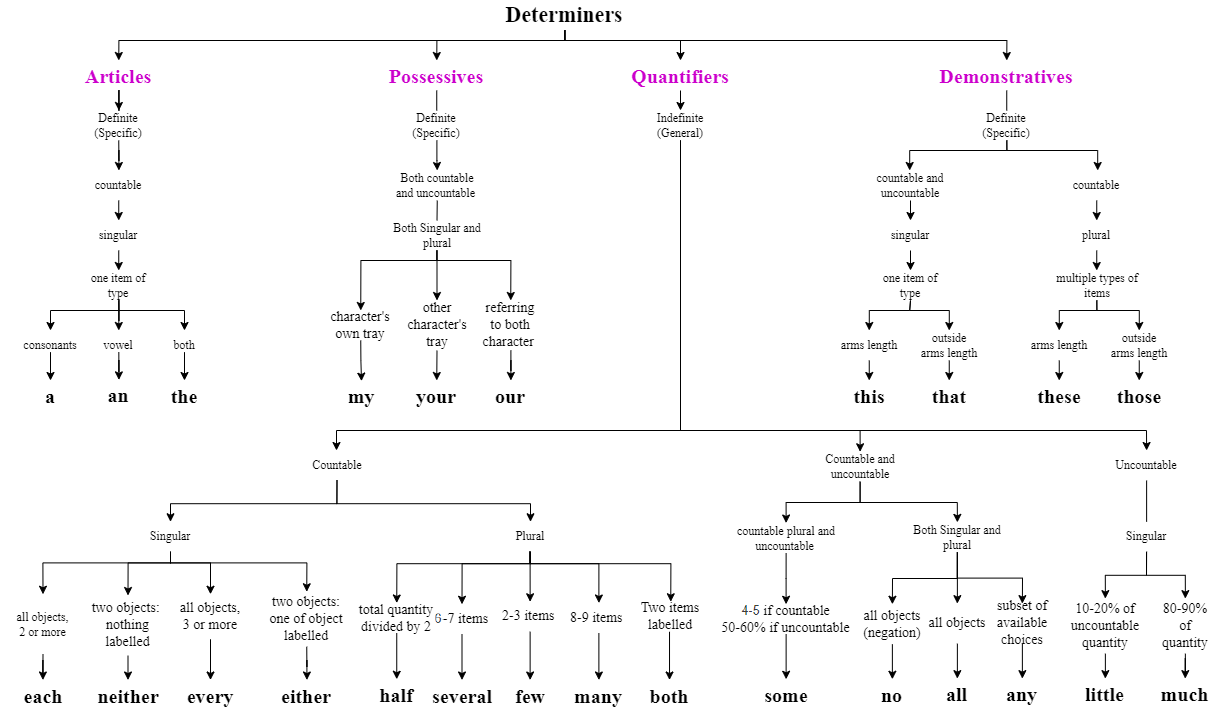}

   \caption{Organization and characteristics of the 25 determiners in DetermiNet.}
   \label{fig:detorg}
\end{figure*}

The task is to choose a set of objects that is consistent with the given \{\textit{determiner noun}\} pair. Examples are ``those apples'' or ``either orange''. Beyond just object detection, the task tests the ability to understand the logical semantics that define various determiners (see Fig.~\ref{fig:detorg}), such as selecting the correct number of requested objects. Simply returning all or random instances of the queried noun would not lead to high performance. Since the focus of DetermiNet is on the logical schema of determiners, high levels of visual realism and diversity are not crucial for benchmarking the ability of algorithms to learn determiners.

Finally, we analyze the performance of SOTA models that were pre-trained to perform visual grounding, so as to see if SOTA deep learning models can learn to understand the logical schema governing determiners.

In summary, our contributions are as follows:
\begin{enumerate}
  \item We developed DetermiNet, the first large-scale diagnostic dataset covering the \textit{determiners} word class, with 250,000 examples across 25 determiners from all four main types of determiners (Articles, Possessives, Demonstratives and Quantifiers). 
  \item We show that the core task of learning determiners is very challenging -- even an oracle model struggles to learn the determiner schema from a few hundred examples and requires a large dataset.
  \item We find that state-of-the-art visually-grounded models show only moderate results on DetermiNet, hence much more work is needed to perform well on the end-to-end task.
\end{enumerate}

\section{Related work}
\subsection{Datasets}
There has been substantial work in developing datasets for visual question answering and referring expressions. However, referring expression datasets which include egocentric points of view and focus on the full coverage of the determiner class for referring is limited (see Table~\ref{table:datasets}). While a dataset like Flickr30k Entities \cite{plummer2015flickr30k} contains some determiners, its coverage is narrow, with only 5.33\% being non-articles. Furthermore, the captions do not consistently capture the semantics of the determiner. For example, although one particular caption specifies ``some people ...'', all the people (\ie many) are labelled instead of just a relatively small number of people. Lastly, Flickr30k Entities is used as a phrase grounding dataset rather than a referring expression dataset, hence it is excluded from Table~\ref{table:datasets}.


\begin{table}[h!]
    \caption{Comparison of datasets for referring expressions \cite{islamcaesar}. A, P, D, Q, Exo and Ego stand for Articles, Possessives, Demonstratives, Quantifiers, Exocentric and Egocentric respectively.}
    \resizebox{\columnwidth}{!}{
    \begin{tabular}{l|c|c|c|c|c|c|c}
    
    \textbf{Datasets} & \textbf{A} & \textbf{P} & \textbf{D} & \textbf{Q} & \textbf{View} & \textbf{Images}  &  \textbf{Type} \\
    \hline
         RefCOCO \cite{kazemzadeh2014referitgame}  & Y & N & N & N & Exo & 19,994 & Real\\
         RefCOCO+ \cite{kazemzadeh2014referitgame}  & Y & N & N & N & Exo & 19,992 & Real \\
         RefCOCOg \cite{mao2016generation} & Y & N & N & N & Exo & 26,711  & Real\\
         CLEVR-Ref+ \cite{liu2019clevr} & Y & N & N & N & Exo & 99,992 & Synth\\
         YouRefIt \cite{chen2021yourefit} & Y & N & N & N & Exo & 497,348 & Real \\
         \hline
         DetermiNet & Y &  \textcolor{blue}{\textbf{Y}} &  \textcolor{blue}{\textbf{Y}} &  \textcolor{blue}{\textbf{Y}} & \textcolor{blue}{\textbf{Ego}} & 250,000 & Synth \\
    \end{tabular}
    }
    \label{table:datasets}
\end{table}

\subsection {Tasks}

A greater confluence between computer vision and natural language processing research has given rise to increasingly complex mixed-modality tasks such as Visual Question Answering (VQA) \cite{antol2015vqa, johnson2017clevr, shridhar2020alfred} and Referring Expression Comprehension (REC) \cite{kazemzadeh2014referitgame, mao2016generation, thomason2022language}. 

Both the datasets for VQA and REC are similar in that the input comprises of images or videos, and a language query is given as a caption. For VQA tasks, the model has to respond to the query by classifying the correct answer out of several potential choices. REC tasks are considered to be a harder problem as the model has to respond by predicting the bounding box coordinates or segmentation masks that identify the object of interest. Nevertheless, both tasks require a combined understanding of language attributes such as colour, shape and size, and visual attributes such as of object classes and location.

The DetermiNet dataset is related to the REC task, where the model needs to identify the object of interest either using bounding boxes or segmentation masks. However, our task defers from existing REC tasks in two ways. 

Firstly, DetermiNet's captions involve only two components, a determiner followed by a noun, instead of descriptive adjectives such as colours or shapes \cite{johnson2017clevr}, or other nouns such as people or objects \cite{kazemzadeh2014referitgame}. This forces models to learn and reason using a new word class, instead of using visual features and spatial representations pre-learned from other visual datasets. 

Secondly, REC tasks usually tests the identification of a single object. However, DetermiNet requires models to predict multiple objects based on the query given, instead of identifying only a single instance. For example, if an image has three apples and two carrots and the query is ``all apples'', the model needs to predict all three bounding boxes instead of a single one. This is the biggest difference between DetermiNet and other REC tasks. DetermiNet allows the development of models to identify multiple objects that correspond to the determiner schema. 

Since DetermiNet allows multiple solutions to be proposed, there can also be multiple combinations of possible solutions. For example, given the same image with three apples and two carrots, and the query ``any apples'', the total number of correct solutions quickly increases to $C(3,1) + C(3,2) + C(3,3) = 7$. The task evaluation metric should not penalise possible solutions and should accommodate the model prediction accordingly. To our knowledge, there are no REC or VQA tasks that support multiple combinations of solutions.

\subsection{Models}
Existing visually grounded models can combine language and visual modalities to achieve superior performance on many downstream tasks such as those in Grounded Language and Visual Question and Answering.



Dual Encoder models such as MDETR~\cite{kamath2021mdetr} and GLIP~\cite{zhang2022glipv2} use an image and text encoder model to encode the inputs before implementing a deep fusion or transformer layer to train the model on the image caption pairs. Seq2Seq models such as OFA~\cite{wang2022ofa} follow the likes of GPT~\cite{radford2018improving} by processing multimodal inputs using byte sequence representation. A unified vocabulary approach to vision and language tasks is taken to perform the grounding tasks. SOTA models such as MDETR and OFA perform really well on visual grounding tasks by achieving 87.5\% AP and 92.0\% AP respectively on the RefCOCO dataset.


However, these models have been largely evaluated against referring expression datasets that are dependent on the spatial and visual attributes of objects. Hence, a more challenging dataset is needed to determine if these SOTA models are robust to solve language and egocentric-based object referencing, like in natural language.

\section{The DetermiNet dataset}

DetermiNet is the first visuo-linguistic dataset based on the determiners word class. Fig.~\ref{fig:detorg} describes our determiner schema that describes which object and how many of those objects should be selected. The dataset was generated synthetically using this schema and focuses on the referencing and quantification of noun phrases. Determiners are largely used from an egocentric perspective, and their properties requires models to perform deeper and more complex reasoning to accomplish the visual grounding task. Careful curation of the dataset was conducted to account for these complexities.

To provide a comprehensive coverage, our dataset includes all four main types of determiners~\cite{jarrodtaylor_2021, leech2014word}, namely:

\noindent 
-- \textbf{Articles:} identify nouns which the speaker is referring to\\
-- \textbf{Possessives:} signify ownership of the noun\\
-- \textbf{Demonstratives:} isolate nouns that are being referred to\\
-- \textbf{Quantifiers:} describe the amount of the referred noun\\

\subsection{Dataset design and construction}
DetermiNet is a synthetically generated dataset based on an end-to end-pipeline developed in Unity. Scene and phrase generations were done through predefined scene configurations based on the scene chart. Since the logic governing determiners is unrelated to the level of visual realism, DetermiNet follows the approach of synthetic data with visual simplicity~\cite{goyal2020rel3d, johnson2017clevr, liu2021learning, yi2019clevrer}. For example, CLEVR~\cite{johnson2017clevr} and CLEVRER~\cite{yi2019clevrer} use only 3 shapes, 2 materials and 8 colors; the background is uniform.

\subsection{Dataset statistics}

DetermiNet has a comprehensive coverage of 25 determiners. We generated 10,000 image-caption pairs per determiner, totaling 250,000 samples. We describe the breakdown of our train, test, and validation splits in Table~\ref{table:statistics}.

\begin{table}[h!]
    \caption{Statistics for train, test and validation splits}
    \resizebox{\columnwidth}{!}{
    \begin{tabular}{l|c|c|c}
    
    \textbf{Splits} & \textbf{Samples} & \textbf{Objects} & \textbf{Ground truth b-boxes} \\
    \hline
         Train & 175000 & 2799790 & 460200\\
         Validation & 25000 & 399654 & 66023 \\
         Test & 50000 & 799756 & 131460 \\
    \end{tabular}
    }
    \label{table:statistics}
\end{table}

\begin{figure}
  \centering
   \includegraphics[width=\linewidth]{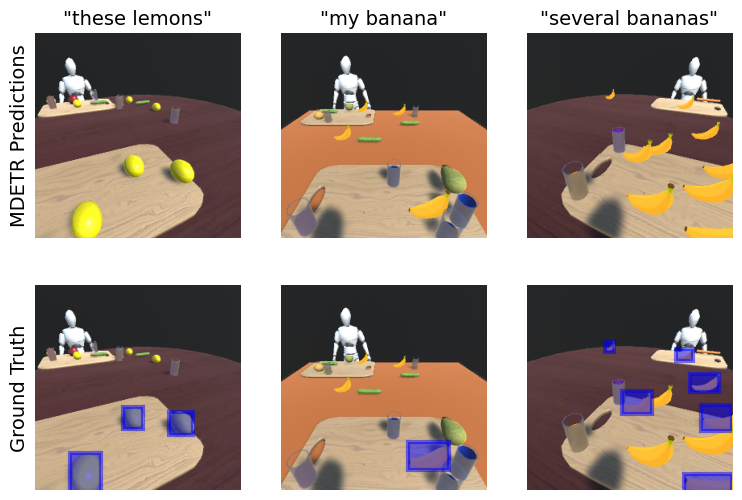}

   \caption{Examples from DetermiNet, with image, phrase, target bounding boxes and segmentation masks shown.}
   \label{fig:exdet}
\end{figure}

In total, our dataset includes a variation of 15 object classes, including 5 countables starting with consonant sounds (\eg ``a lemon''), 5 countables starting with vowel sounds (\eg. ``an apple'') as well as 5 uncountable substances (\eg. ``some grape juice''). Ground truths are determined by the object which the determiner is referring to. This referred object will then be labelled as part of the ground truth annotations (Fig. \ref{fig:exdet}). Variations indicate the number of different permutations of the object, while the number of objects spawned indicate the possible count of that particular item spawned in the scene. A summary of the scene and object variations is shown in Table~\ref{table:variations}. 

\begin{table}[h!]
    \caption{Scene variations}
    \resizebox{\columnwidth}{!}{
        \begin{tabular}{l|c|c}
        \textbf{Object} & \textbf{Variations} & \textbf{No. spawned in scene} \\
        \hline
             \textbf{\textit{Referred}} objects & 15 & 1-9\\
             All objects & 15 & 10-20 \\
             \hspace{3mm} Countables (consonant) & 5 & 1-20 \\
             \hspace{3mm} Countables (vowels) &  5 & 1-20 \\
             \hspace{3mm} Uncountables &  5 & 1-20 \\
             Trays & 2 & 2 \\
             Tables & 2 & 2 \\
             Tray positions & 3 & - \\
             Camera positions & 3 & - \\
        \hline
        \end{tabular}
    }
    \label{table:variations}
\end{table}

\subsection{Scene generation and ground truth annotation}
\label{sssec:gt}

DetermiNet is based on the interaction of two avatars at a table. We randomly spawn the positions of objects, as well as generate different perspectives. Configuration parameters were used to determine the construction of each scene, providing a unified interface for scene generation. These configuration parameters follow the tree in Fig.~\ref{fig:detorg}, and can be adjusted to the user's own definitions. Attributes include type of object (countability), number of referred objects (plurality), spawn locations and distance from the viewer. Egocentric viewpoints of the viewer were generated by attaching the camera to the viewer's head and directing the camera to focus towards the center of the table. We varied the avatars' positions to generate multiple perspectives.

Images were rendered using Unity3D. Camera projections were used to check for visibility of the spawned objects and collision detectors were put in place to ensure that objects did not intersect. Different objects (tray, tables) were also sampled to be used as random spawn locations.

Mesh vertices were projected onto the camera's 2D space to extract bounding boxes for all objects, modeling a perfect object detector. Unity's Image Synthesis module was used to generate object segmentation masks.

\subsection{Phrase generation}

DetermiNet uses the task context of ``pass me \{\textit{determiner noun}\}'', \eg. ``pass me an apple'', ``pass me that apple''. For simplicity, we omitted ``pass me''. Hence, the phrases are simple captions with only a determiner and its noun phrase (Fig. \ref{fig:exdet}), \eg. ``an apple'', ``this apple'', ``some grape juice''. Additionally, we follow this phrasing format while keeping errors in grammatical structure minimal. For example, ``pass me all apples'' is sufficient to capture the task instead of ``pass me all the apples''.

\subsection{Evaluation metric for DetermiNet}

Since the task is to evaluate bounding box predictions, we used the detection evaluation metric used by COCO, specifically the average precision (AP) metric with IoU thresholds ranging from 0.5 to 0.95.

The DetermiNet dataset contains scenarios where different combinations of solutions can be correct. For instance, for an image with three apples and a query specifying ``an apple'', there are three equally correct solutions. However, a correct bounding box prediction should only contain one bounding box instead of three. If all three bounding boxes are predicted, the evaluation metric should evaluate the prediction as one true positive and two false positives. 

To account for multiple correct solutions during evaluation, we developed a ground truth correction function that compares the model's predicted bounding boxes against all the relevant bounding boxes that satisfy both the determiner and noun conditions. The function chooses the ground truth bounding box that has the highest IoU with the predicted bounding box, and discards the rest of the relevant ground truth bounding boxes based on the quantity specified by the determiner. 

The modified ground truth annotations are then used to evaluate the predictions. This way, if a model predicts three bounding boxes instead of one, the prediction with the highest IoU and prediction score will be treated as true positive, and the other two predictions treated as false positive.

\section{Experiments}
In this section, we verify the challenge posed by the dataset to refer or quantify objects of interest using five models. Since the DetermiNet task is similar to the REC task, models need to predict bounding boxes which were evaluated using the Average Precision (AP) evaluation metric. Before evaluation, the ground truth bounding box annotations were modified to account for multiple combinations of correct solutions.

\begin{figure*}
  \centering
   \includegraphics[width=\linewidth]{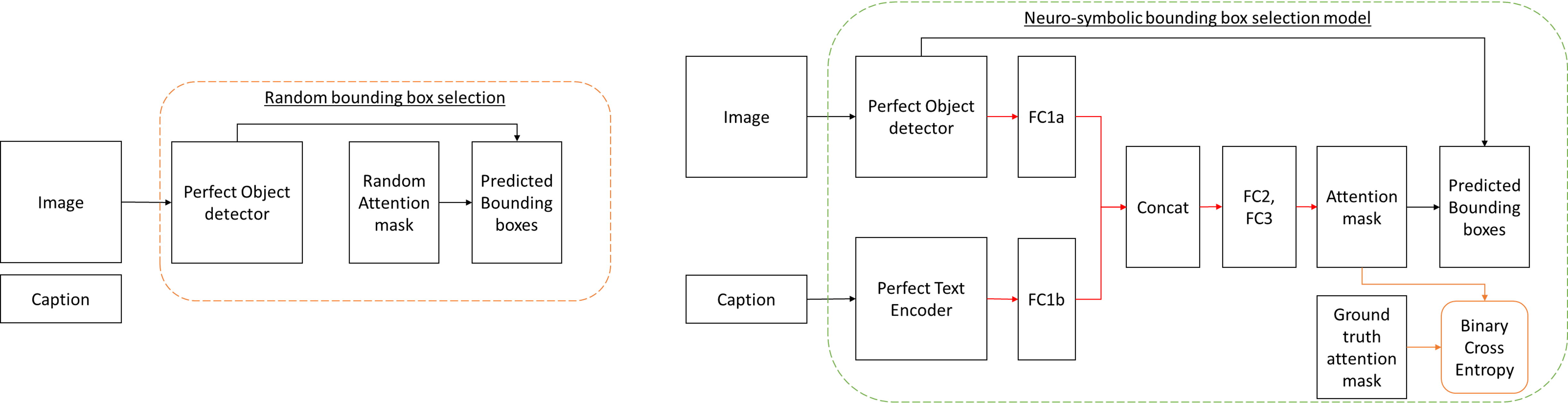}

   \caption{Random and neuro-symbolic model architectures. Weights of fully connected (FC) layers were optimized by backpropagation.}
   \label{fig:modelarch}
\end{figure*}

\subsection{Random selection model}

The first model is a random bounding box selection model (Fig.~\ref{fig:modelarch}). This model has two components. The first is a perfect object detector (see~\ref{sssec:gt}) that tags all objects with class labels and their corresponding bounding boxes. 

The second component sampled prediction scores between 0 to 1 from a uniform distribution and generated positive and negative masks based on a threshold of 0.5 which was used to select bounding boxes as predictions. 

In short, the perfect object detector generated a list of bounding boxes and the attention mask randomly selected a subset of bounding boxes as predictions without using information of either determiner or noun.

\subsection{Neuro-Symbolic oracle model}
The neuro-symbolic model (Fig.~\ref{fig:modelarch}) was developed to isolate the main challenge of the dataset, which is to classify objects of interest based on the concept specified by the determiner. Hence, this model tackles the DetermiNet dataset as a classification problem, similar to VQA models. 

Like the random selection model, a perfect object detector was used to identify all the object bounding boxes, class labels and volume of liquid within the object. These three information were fed to a single feedforward layer with 128 units to embed the visual information. 

A perfect text encoder converted the two-part caption specifying the determiner and the noun into two one-hot encoded vectors. The first one-hot vector of length 25 represented the determiner, and the second one-hot vector of length 16 represented the noun. The two vectors were concatenated and fed to another feedforward layer with 128 units to embed textual information. 

The output of the two embedding layers were concatenated and fed to two feedforward layers, each with 256 units, followed by a final classification layer with sigmoid activation function. 

A ground truth attention mask was generated by comparing all the objects detected in the image against the ground truth bounding boxes such that masking the list of object bounding boxes detected by the perfect object detector will provide the ground truth bounding boxes. The model was trained to predict the ground truth attention mask using binary cross entropy for 30 epochs. 

The model's prediction scores from the classification layer and bounding boxes extracted by the perfect object detector were used for evaluation. The neuro-symbolic model can be considered to be an \textbf{oracle model}, as it received ground-truth information about all the objects in the image, and it only needs to learn to predict the correct bounding boxes given the determiner and noun.

\subsection{SOTA deep learning models}
To verify the full challenge posed by DetermiNet, we fine-tuned three SOTA visual grounding models, OFA\cite{wang2022ofa} with ResNet-152 backbone, GLIP \cite{zhang2022glipv2} and MDETR \cite{kamath2021mdetr} with ResNet-101 backbone for 5 epochs on our dataset. 

OFA's weights were pretrained on RefCOCO and VG datasets, GLIP's weights were pretrained on O365, GoldC, CC3M and SBU datasets while MDETR's weights were pretrained on the RefCOCO, VG and Flickr datasets. Both image and captions were passed as inputs to the SOTA models, and the bounding box predictions were obtained as outputs. The object class prediction was not relevant to our DetermiNet task, so we set category ID to 1 for all predictions. While GLIP and MDETR models returned multiple bounding box predictions and scores, OFA is designed to predict only one bounding box per image.

\section{Benchmarking models on DetermiNet}
After correcting the ground truth annotations to account for multiple solutions, the random bounding box selection model demonstrates the worst performance of 9.8\% AP. Even though the random model has the perfect object detection module, randomly selecting different quantities of different objects without considering the textual information leads to poor performance. This can be treated as the lower-bound performance for the DetermiNet dataset.

In contrast, the oracle demonstrates the highest performance of 93.5\% AP (Table~\ref{table:modelperf}) as it receives object class and textual information while only needing to learn the determiner schema. Since the oracle model is only tested on semantics to provide a \textbf{rough upper-bound} for DetermiNet, its performance \textbf{should not be directly compared} against end-to-end models which learn both object detection and determiner semantics, and whose learning performance is difficult to disentangle. When the oracle uses MDETR object detection outputs instead of perfect detection, overall AP fell to 62.8\%.


\begin{table}[H]
\begin{center}
\caption{Model performance after correcting ground truth annotations. *OFA only predicts one bounding box.}
\label{table:modelperf}    
\begin{tabular}{l|c}
\textbf{Models} & \textbf{AP@IoU=0.5:0.95} \\
\hline
     Random                        &  9.8 \\
     Oracle                & 93.5 \\
     \hline
     OFA \cite{wang2022ofa}        & 20.6*  \\
     GLIP \cite{zhang2022glipv2} & 55.0 \\
     MDETR \cite{kamath2021mdetr}  & 70.6 \\

\hline
\end{tabular}
\end{center}
\end{table}

\begin{figure}[H]
  \centering
   \includegraphics[width=\linewidth]{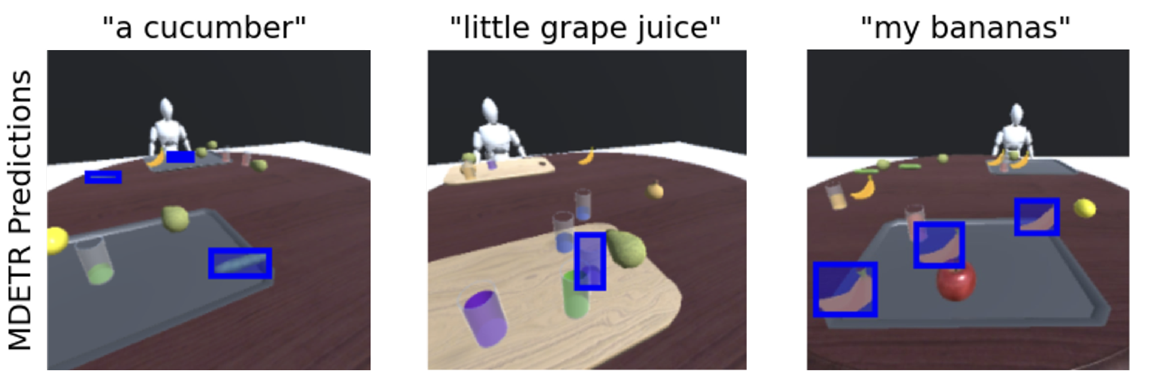}

   \caption{MDETR suffers from high false positives as bounding box predictions (blue) are not constrained (left) but shows learning of uncountable quantifiers (middle) and possessives (right).}
   \label{fig:mdetrexp}
\end{figure}

When comparing end-to-end finetuned models, OFA performs the worst, as it is only able to predict one bounding box, similar to the REC task condition, contributing to high false negatives. GLIP achieves 55.0\% while MDETR achieves the best performance of 70.6\% AP (Table \ref{table:modelperf}). Although MDETR's bounding box predictions are impressive to identify the reference objects, the model does not constrain its predictions according to the determiners schema, incurring high false positive predictions. Conversely, MDETR performs well on uncountable quantifiers and possessives (Fig.~\ref{fig:mdetrexp}). This is likely because MDETR gets the raw RGB image as input, allowing it to understand and reason about volume levels within a cup or the presence of the referred object on the tray. 

Refer to the \textbf{Supplementary Material} for the confusion matrix broken down for each determiner and class, as well as the performance when constraining each model's predictions to just the top-1 bounding box (similar to OFA's constraint).

\subsection{Embedding of determiners}

\begin{figure*}
  \centering
   \includegraphics[width=\linewidth]{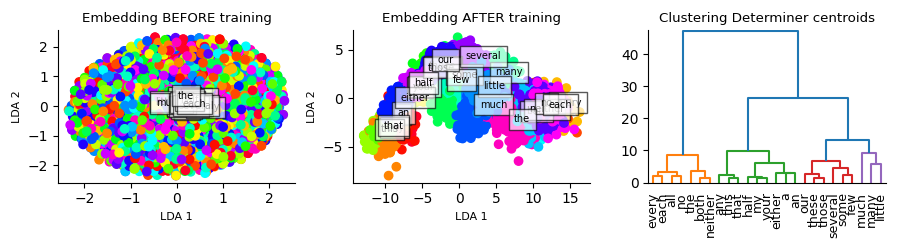}

   \caption{Clustering 25 determiners represented in the last feature layer of the oracle model using LDA.}
   \label{fig:tsne}
\end{figure*}

To study how the dataset is represented in both an untrained and trained network, we extracted the neural activity of the layer before the attention mask classifier. The neural activity was clustered using Linear Discriminent Analysis, with the determiner labels as targets. Before training, the neural representations corresponding to 25 determiners were highly overlapped and the centroid coordinates for each determiner class occupied the same space (Fig.~\ref{fig:tsne}, left). 

As training progressed, the embedding of the 25 determiners evolved into clusters (Fig.~\ref{fig:tsne}, middle). The dendrogram (Fig.~\ref{fig:tsne}, right) represents the euclidean distance between centroids after training. With training, the network learns a representation that seemingly corresponds to the organization of determiners in Figure \ref{fig:detorg}.

Neural representations for ``a'' and ``an'' occupy the same subspace as they obey the same articles determiner schema. We can see similar clustering of determiner subclasses such as ``both'' and ``neither'' which fall under quantifiers and ``this'' and ``that'' which fall under demonstratives. However, some determiners such as ``the'' and ``our'' do not occupy the same subspaces as articles or possessives, suggesting that the model struggles to disentangle them.
Surprisingly, unlike the oracle model, text encoders in established vision-language models such as CLIP~\cite{radford2021learning} and BLIP-2~\cite{li2023blip} do not demonstrate distinct organization of determiners \textbf{(see Supplementary Material).}

\subsection{Ablation study}
To determine the importance of determiners and nouns in the DetermiNet task, we conducted ablation studies using oracle and MDETR models where the determiner, noun or both determiner and noun were masked during evaluation. 

\begin{table}[h!]
\begin{center}
\caption{Ablation study with masked captions. Performance reported AP@IoU=0.5:0.95}
\label{tab:ablate}    
\begin{tabular}{c| c |c}
\textbf{Ablation condition} & \textbf{Oracle} & \textbf{MDETR} \\
\hline
     Noun+ / Det+   & 93.5 & 70.6  \\
     Noun+ / Det--  & 71.3 & 56.3 \\
     Noun-- / Det+  & 11.3 & 11.3 \\
     Noun-- / Det-- & 9.8 & 0.2 \\
 \hline
\end{tabular}
\end{center}
\end{table}

Masking determiners while feeding in the noun is similar to a query-based object detection task. The decrease in performance for the oracle model was 22.2\% while MDETR suffered a decrease of  14.3\%, suggesting that MDETR learnt to predict bounding boxes using most of the determiner concepts, though not as well as the oracle model. 

When the determiner was given but the noun was masked, AP dropped significantly since the object to be identified was not known. Finally, when both determiner and noun were omitted, the oracle performed similarly to the lower bound random model while MDETR performed much worse since it also had to perform object detection.

Nevertheless, SOTA models do learn some determiner concepts, and lower performance can be attributed to errors in both object detection and bounding box classification.

\subsection{Dataset efficiency}
Since 10,000 examples per determiner in the full dataset is presumably way beyond what humans require to learn determiners well, we trained the oracle and MDETR models on randomly sampled subsets (N=6) of DetermiNet training samples to determine how much data is needed for the models to learn the determiner schema. 



\begin{table}[h!]
\centering
\caption{AP@IoU=0.5:0.95 with standard deviation attained after training models on 10, 100, 1000 samples per determiner. }
\label{tab:dataeff}    
\begin{tabular}{l|c |c |c}
\textbf{Samples} & \textbf{10} & \textbf{100} & \textbf{1000} \\
    \hline
     Oracle                   &  17.9$\pm$0.6  & 29.6$\pm$0.4 & 44.7$\pm$3.3 \\
      \hline
     MDETR        & 2.8$\pm$1.0 & 33.5$\pm$1.2 & 55.0$\pm$0.8  \\

\hline
\end{tabular}
\end{table}

Since the oracle has a perfect object detector and text encoder, the increase in oracle performance is attributed solely to the learning of determiner schema. Despite the isolation of training, the oracle model struggles to learn the concept of determiners even with 1,000 examples per determiner. This could be because the oracle model has 188,308 trainable parameters and a large dataset is needed to optimize the weights accordingly. Conversely, MDETR has 185 million parameters but was pre-trained to perform object detection. After fine-tuning MDETR with 1,000 examples per determiner, its performance matches the ablation condition where the model can achieve 56.3\% without needing to learn determiners (Table~\ref{tab:ablate}), suggesting that the faster improvement is likely due to improved object detection in DetermiNet, rather than learning about determiners. Nevertheless, DetermiNet follows a scaling law that is consistent with other visual recognition tasks.

\section{Transfer of learning to real images}
We curated a dataset with 100 real world images and captions using images from COCO~\cite{chen2015microsoft}. The oracle model achieved decent zero-shot performance on the real-image samples (78.1\%), demonstrating a neural network's ability to generalize to real images if object detection works well.

\begin{figure}[h!]
  \centering
   \includegraphics[width=\linewidth]{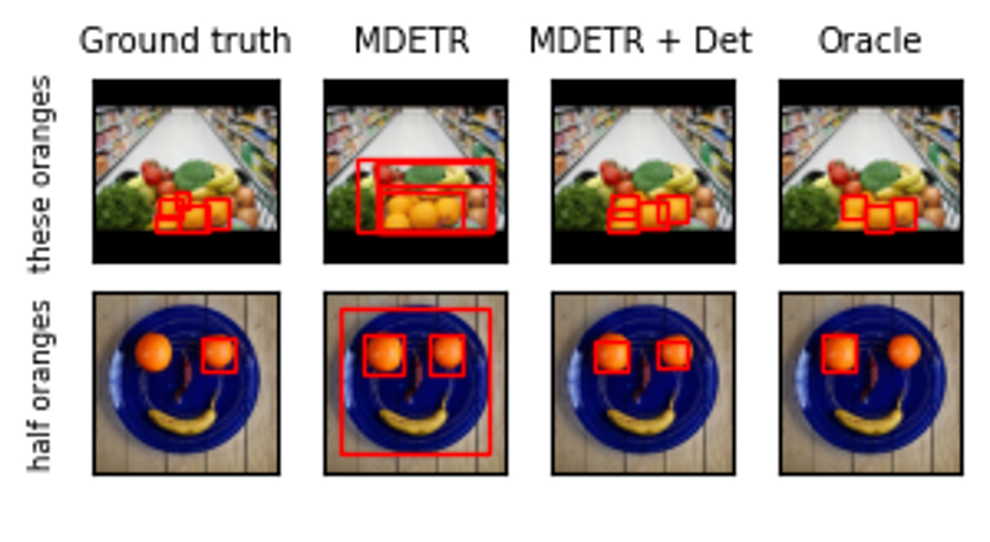}

   \caption{Ground truth, pretrained MDETR, MDETR fine-tuned on DetermiNet and Oracle model predictions on 100 real images.}
   \label{fig:realpred}
\end{figure}

Although MDETR was pre-trained on RefCOCO, it struggled to refer and quantify individual objects according to the determiner schema (10.4\%) since RefCOCO did not account for such determiner concepts (Table~\ref{table:datasets}) and instead predicted single bounding boxes for a collection of objects (Fig.~\ref{fig:realpred}).  Fine-tuning MDETR on the synthetic DetermiNet significantly increased performance to 19.5\% as the model learned to identify and quantify each object (Fig.~\ref{fig:realpred}, top row), suggesting that the determiner concepts learned from the synthetic dataset transferred to real images to a certain extent. However, MDETR still struggles with some determiner concepts such as ``half'' (Fig.~\ref{fig:realpred}, bottom row). The far lower MDETR performance could be due to poor object detection, separate from learning the semantics of determiners. The real-image test samples will be made available along with the synthetic DetermiNet.

\begin{table}[h!]
\centering
\caption{Zero-shot evaluation on real-image dataset}
\label{tab:realmodelperf}    
\begin{tabular}{l|c}
\textbf{Models (Tasks pretrained on)} & \textbf{AP@IoU=0.5:0.95} \\
\hline
     Oracle                   &  78.1 \\
          \hline
     MDETR (Pretrained)                    &  10.4 \\
     MDETR (Finetuned on DetermiNet)       &   19.5 \\
\hline
\end{tabular}
\end{table}

\section{Current limitations}

Since the dataset focuses on referencing and quantification, we omitted the use of wh-determiners (\eg. ``where'', ``what''), which are mainly used in question answering tasks. Since we constrained our captions to fit the task context of ``pass me \{determiner, noun\}'', comparison determiners such as ``more'' and  ``less'' were left out for now, as they require multiple sets of nouns. Furthermore, gender-specific possessives such as ``his'' and ``her'' were omitted, as gender recognition is not the focus of this work. Additionally, the composition of multiple determiners (\eg. ``pass me \underline{some} of \underline{those} apples) will be explored in future work. 

Parameter-efficient finetuning, or adding the semantic module of the oracle to a trained detector such as MDETR can serve as an additional evaluation to disentangle the learning performances of object detection and determiner semantics in end-to-end models.


\section{Conclusion}
We present the DetermiNet dataset to determine if models can learn object referencing and quantification for all four major determiner categories. The dataset accommodates multiple combinations of possible solutions, as in a natural language context. Since the dataset images and ground truth annotations were synthetically generated, it allows for rapid reconfiguration of parameters, scenes and object classes to increase the challenge posed by the dataset. 

Our experiments demonstrate that although state-of-the-art visual grounding models are able to identify objects of interest, they do not perform well on the overall task. While they can learn the semantics of some determiners and transfer the concept to real images, they require exponential amounts of data to learn the determiner schema and struggle to handle ambiguity when considering multiple combinations of possible solutions. 


In summary, DetermiNet highlights determiners as important and complex but neglected, and formulates a common task framework for all 4 determiners types. It shows the current limitations of visual grounding models in learning determiner schemas in referencing and quantification. Good oracle results on real images suggests the “determiner logic module” could be used for captioning, VQA, etc.


\section{Acknowledgements}
This was supported by an A*STAR CRF award (C.T.), and by the Design and Artificial Intelligence department, Singapore University of Technology and Design (C.L.).

\pagebreak

{\small
\bibliographystyle{ieee_fullname}
\bibliography{egbib}
}
\clearpage

\end{document}